\documentclass{article}

\usepackage{xcolor}

\usepackage[preprint]{corl_2026} 

\usepackage{graphicx}
\usepackage{subcaption}
\usepackage{amsmath}
\usepackage{amssymb}
\usepackage{booktabs}
\usepackage{algorithm}
\usepackage{algorithmic}
\usepackage{tikz}

\graphicspath{{figs/}}

\title{Actuator Reality Shaping \\ for Zero-Shot Sim-to-Real Robot Learning}

\author{
  \textbf{Satoshi Yamamori$^{1,2}$, Koji Ishihara$^{2}$, Kenjiro Minamikawa$^{1}$, Ryosei Ohmori$^{1}$,}\\
  \textbf{Taiyo Yasaki$^{1}$, Norikazu Sugimoto$^{2}$, and Jun Morimoto$^{1, 2}$}\\
  $^{1}$Graduate School of Informatics, Kyoto University, Kyoto, Japan\\
  $^{2}$Dept. of Brain Robot Interface, Computational Neuroscience Labs, ATR, Kyoto, Japan\\
}

\begin{document}

\maketitle

\begin{abstract}
Sim-to-real transfer in robot learning is often limited by discrepancies between the ideal actuator dynamics assumed during policy training and the nonlinear, hardware-dependent behavior of physical motors. While conventional approaches attempt to bridge this gap by increasing simulator fidelity through system identification, domain randomization, or learned actuator models, we introduce an alternative paradigm: actuator reality shaping. Instead of modifying the simulator to match the real world, our method shapes the closed-loop behavior of physical actuators to match the idealized second-order reference dynamics used in simulation. By equipping each joint with a two-degree-of-freedom feedforward--feedback controller, we decouple reference-response shaping from robust stabilization, thereby providing a standardized actuator interface for reinforcement learning policies. As a result, policies trained only with the prescribed reference model can be deployed zero-shot on real hardware without task-level fine-tuning or learned actuator models. We validate the approach on a single-joint high-gear-ratio servo under external loads and a 7-DOF robotic arm reaching task, where actuator reality shaping substantially reduces sim-to-real tracking error and improves zero-shot task performance compared with standard servo-control and representative real-to-sim-to-real baselines. We further demonstrate zero-shot transfer on a wheeled-legged robot driving over a slope and a humanoid robot walking, suggesting that actuator reality shaping can serve as a reusable interface for robot learning across diverse hardware platforms. 
\href{https://syamamori.github.io/ActuatorRealityShaping.github.io/}{Project page}
\end{abstract}


\section{Introduction}

Reinforcement learning (RL) has enabled impressive locomotion and manipulation
controllers when trained in GPU-accelerated, large-scale physical
simulation~\citep{hwangboLearningAgileRobust2019,
kumarRMArapidMotorAdaptation2021,schulmanProximalPolicyOptimization2017,
grandiaDesignControlBipedal2024,fuHumanPlusHumanoidShadowing2024}.
However, bridging the \emph{sim-to-real gap}, the performance degradation that
arises when policies trained in simulation are deployed on physical
hardware, remains a central challenge in robot learning.
A major source of this gap lies at the actuator level.
While rigid-body simulators often model joints as ideal torque sources or
simplified servo systems, real actuators exhibit friction, stiction, backlash,
transmission compliance, inertia mismatch, saturation, and electromechanical
delays.
These unmodeled effects distort the relationship between commanded and realized
joint motion, causing policies optimized in simulation to drive the physical
robot along unintended trajectories.
The resulting mismatch can degrade task performance, amplify tracking errors,
and, in dynamic tasks, destabilize the robot.

The dominant strategy for reducing the sim-to-real gap has been to make the
simulator more realistic~\citep{tobinDomainRandomizationTransferring2017}.
Domain randomization (DR) trains policies over randomized physical parameters so
that the real system is likely to lie within the training
distribution~\citep{tobinDomainRandomizationTransferring2017,akkayaSolvingRubiksCube2019}.
Although effective in many settings, this robustness often comes at the cost of
task-specific performance, since policies must succeed across a broad range of
dynamics rather than exploit a precise actuator model.
A complementary line of work augments the simulator with physics-based or learned
actuator models identified from input--output
data~\citep{hwangboLearningAgileRobust2019,bjelonicBridgingGapSystematic2025},
or learns online adaptation modules that infer latent environment or actuator
parameters during deployment~\citep{kumarRMArapidMotorAdaptation2021}.
Despite their success, these approaches share a common dependency: they adapt the
simulation or policy to a particular hardware distribution.
As a result, they typically require hardware-specific data collection or
identification, which must be repeated when actuators are replaced, degrade over
time, or are transferred to a different robot platform.
This dependency limits the reusability of policies trained in simulation and
motivates an alternative interface that standardizes the actuator behavior seen by
the policy, rather than continually adapting the simulator to each hardware
instance.

\begin{figure}[t]
  \centering
  \includegraphics[width=0.99\linewidth]{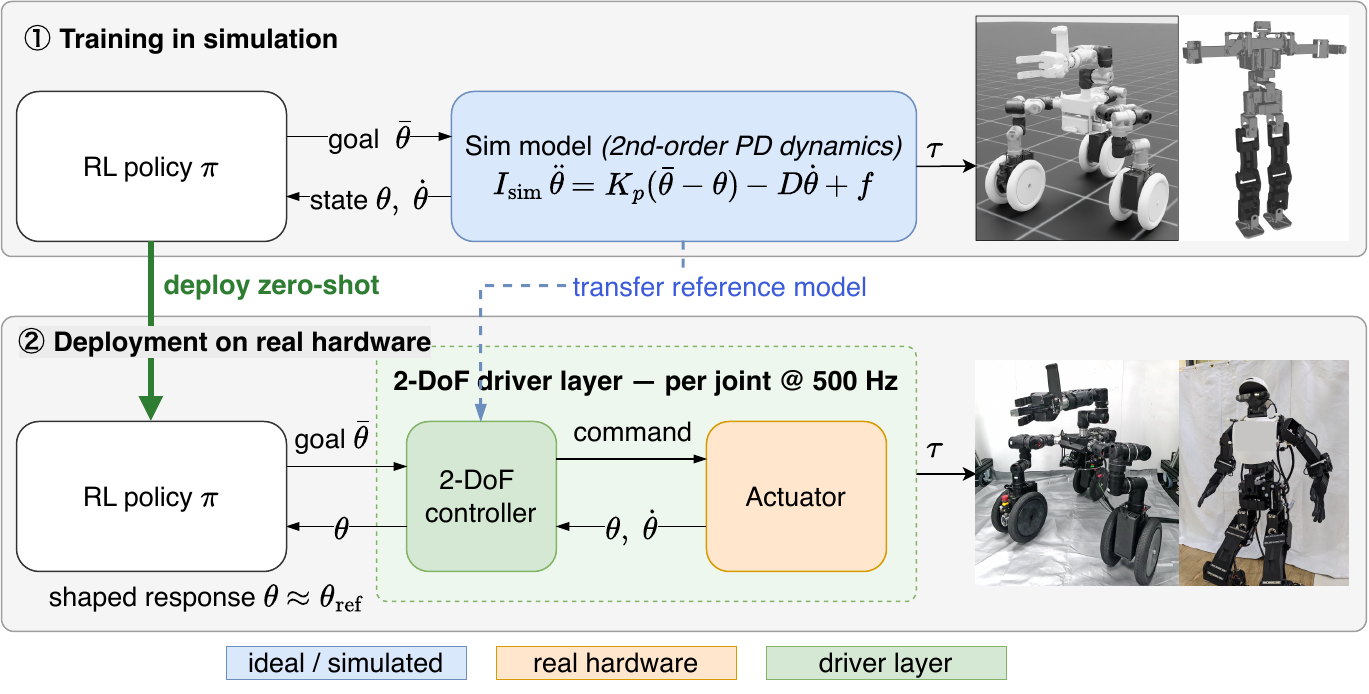}
  \caption{\textbf{Actuator reality shaping.} We make the real actuator behave
    like the simulator's ideal model, rather than the reverse.
    \emph{(Top)} simulation with the ideal second-order reference model;
    \emph{(Bottom)} zero-shot hardware deployment, where a per-joint 2-DoF driver
    shapes the response so that $\theta \approx \theta_{\mathrm{ref}}$.}
  \label{fig:concept}
\end{figure}

We propose an alternative perspective: \emph{actuator reality shaping}. Instead of modifying the simulator to match each physical actuator, we modify the low-level actuator controller so that the hardware presents to the policy the same reference dynamics used during simulation. As illustrated in Figure~\ref{fig:concept}, an RL policy is trained in simulation against an idealized second-order actuator model, and is then deployed zero-shot on hardware, where a per-joint 2-DoF driver shapes the physical actuator response to match the prescribed reference dynamics. Concretely, we introduce a \emph{two-degree-of-freedom (2-DoF) controller} at each joint driver. Here, ``two-degree-of-freedom'' is a control-theoretic term ~\citep{sugieGeneralSolutionRobust1986}: it denotes a controller with two independent design freedoms, typically realized as a feedforward path and a feedback path. This structure allows the reference-response behavior and robust stabilization against disturbances and modeling errors to be designed separately. The feedforward path specifies the desired closed-loop dynamics seen in simulation, while the feedback path compensates for disturbances and residual actuator-model mismatch. We detail these two paths in Section~\ref{sec:method} and Figure~\ref{fig:system}. The resulting driver layer encourages the real joint to follow the same second-order reference dynamics assumed during policy training, thereby providing a standardized actuator interface between the RL policy and the physical hardware. Consequently, a policy trained against this reference model in simulation can be deployed on the real system without task-level fine-tuning or learned actuator models, as long as the physical actuator can track the prescribed dynamics within its bandwidth and saturation limits.

The resulting 2-DoF layer plays a role analogous to a device driver: it hides hardware-specific actuator behavior behind a common interface exposed to the policy. In our case, this interface is dynamical rather than software-based: the policy interacts with the prescribed second-order reference model used in simulation, while the low-level driver compensates for the physical actuator dynamics. This separation provides three practical advantages. First, it enables \emph{policy-level hardware abstraction}: the same policy can be reused across compatible actuator platforms by redesigning or retuning only the low-level driver, rather than retraining the policy. Second, it provides a \emph{smooth learning interface}: because the policy is trained against a consistent, idealized actuator response, policy optimization is less exposed to hardware-specific nonlinearities such as friction, backlash, and delay. Third, it improves \emph{modularity}: the driver can be designed from a compact actuator description, such as the torque constant, rotor inertia, and bandwidth limits, without requiring task-level real-world rollouts or learned actuator models.

Our main contributions are as follows:
\begin{enumerate} 
\item We introduce the \emph{actuator reality shaping} paradigm for sim-to-real transfer in robot learning, and derive a 2-DoF control architecture that shapes the closed-loop actuator response to match a prescribed simulator reference model. 

\item We validate the proposed architecture through single-joint tracking experiments with controlled disturbance injection, and demonstrate zero-shot sim-to-real transfer on a $7$-DOF robotic-arm reaching task. We compare against the factory cascaded servo controller, a tuned PID controller, and a learned Delta Action baseline (ASAP~\citep{heASAPAligningSimulation2025}). 

\item We further evaluate the generality of the approach through additional zero-shot transfer demonstrations on a wheeled-legged robot with velocity-controlled wheels and a full humanoid robot performing stable forward walking. 

\end{enumerate}

\section{Related Work}
\label{sec:related}

\textbf{Sim-to-Real Transfer in Robot Learning}
Existing sim-to-real methods fall into three groups.
\emph{Domain randomization (DR)}~\citep{tobinDomainRandomizationTransferring2017,pengSimToRealTransferRobotic2018,tanSimToRealLearningAgile2018,andrychowiczLearningDexterousInHand2020,akkayaSolvingRubiksCube2019}
trains under randomized parameters so the real system lies within the training
distribution, but yields conservative worst-case policies and large simulation
budgets; adaptive variants such as SimOpt~\citep{chebotarClosingSimToRealLoop2019}
update the distribution from real rollouts, Rapid Motor Adaptation~\citep{kumarRMArapidMotorAdaptation2021}
distills an online module that infers environment parameters from proprioceptive
history, and ASAP~\citep{heASAPAligningSimulation2025} learns a Delta Action model
from real rollouts to compensate the dynamics mismatch at the policy level.
\emph{Actuator modeling} instead augments the simulator with a model identified
from hardware data---a black-box
network~\citep{hwangboLearningAgileRobust2019,leeLearningQuadrupedalLocomotion2020,mikiLearningRobustPerceptive2022}
or a few interpretable parameters (rotor inertia, friction, damping)---and
implicitly assumes a small residual gap: most demonstrations use
\emph{quasi-direct-drive} actuators with low gear ratios, whose torque response
approximates an ideal current source. High-gear-ratio joints violate this
assumption, as reflected friction, stiction, and backlash dominate the joint
torque and cannot be captured as a simple parametric perturbation.
\emph{Data-driven system identification} estimates environment parameters online
and conditions the controller on them, with roots in classical adaptive
control~\citep{slotineAppliedNonlinearControl1991,astromAdaptiveControl2008}
and biped-locomotion work refining a bootstrap controller against a data-driven
model~\citep{morimotoMinimaxDDP2002,morimotoNonparametricPoincare2009}, or trains
an \emph{environment-aware} policy with an online-identified latent
code~\citep{yuPreparingUnknownUPOSI2017,yuSimToRealBiped2019,pengLearningAgileAnimals2020}.
All these share a structural cost: hardware-specific data and/or online
identification that must be repeated when the hardware changes.
We take the dual perspective---rather than identifying the hardware so the
policy adapts to it, we shape the hardware's closed-loop response at the driver
level so the policy perceives a fixed, simulator-defined plant on every platform;
even high-gear-ratio joints emulate the ideal reference plant, with residual
error absorbed at runtime by the disturbance observer and integral feedback.

\textbf{Two-Degree-of-Freedom and Model Reference Control}
The 2-DoF control architecture decouples reference tracking from disturbance rejection through independent feedforward and feedback designs~\citep{sugieGeneralSolutionRobust1986,araki2dof1984,umenoRobustSpeedControl1991}, and is commonly combined with a disturbance observer to reject friction and load~\citep{ohnishiRobustMotion1987,sariyildizAnalysisDisturbanceObserver2015,chenDisturbanceObserverBasedControl2016}.
However, these approaches operate at the level of individual joint control and are therefore insufficient for generating whole-body motion.
As a result, their use as a \emph{sim-to-real interface layer} has not been sufficiently explored.
Our design shares the reference-model layer idea with a fixed-gain 2-DoF controller derived from a minimal actuator model, avoiding convergence monitoring, gain scheduling, and adaptation--saturation stability issues, and requiring no hardware adaptation data.

\section{Problem Formulation}
\label{sec:problem}

\textbf{Markov Decision Process in Actuator}
We consider a Multi-DoF robot with $n$ revolute joints actuated by DC motors.
An RL policy $\pi: \mathcal{O} \to \mathcal{A}$ maps an observation
$o_t \in \mathcal{O}$ to a goal joint angle $\bar{\theta}_t \in \mathcal{A}$.
In simulation, joint $j$ is governed by a PD controller acting on the goal joint angle:
\begin{equation}
    I_{\mathrm{sim}} \ddot{\theta}_j = K_p (\bar{\theta}_j - \theta_j)
    - D \dot{\theta}_j + f_j,
    \label{eq:sim_dynamics}
\end{equation}
where $I_{\mathrm{sim}}$, $K_p$, $D$ are the simulated inertia, proportional
gain, and damping coefficient, and $f_j$ is the net external joint torque.
The transition operator of the MDP is defined by the dynamics (\ref{eq:sim_dynamics}),
and the policy $\pi$ is optimized against this operator using
RL algorithm, e.g. PPO~\citep{schulmanProximalPolicyOptimization2017}.

\textbf{Actuator Gap}
In simulation, (\ref{eq:sim_dynamics}) implicitly assumes that the commanded
torque $\tau_{\mathrm{cmd}} = K_p (\bar{\theta}_j - \theta_j) - D \dot{\theta}_j$
is produced instantaneously at the joint.
On real hardware, joint $j$ is driven by a motor with dynamics:
\begin{equation}
    J_{\mathrm{real}} \ddot{\theta}_j = K_{\tau i}(i_j - i_{d,j})
    - d_{\mathrm{real}} \dot{\theta}_j - \tau_{f,j}(\dot{\theta}_j)
    - \tau_{\mathrm{ext},j},
    \label{eq:real_dynamics}
\end{equation}
where $J_{\mathrm{real}}$ is the actual rotor inertia (including gear reflection),
$K_{\tau i}$ the torque constant, $i_j$ the current command, $i_{d,j}$ a
disturbance current (e.g., cogging), $d_{\mathrm{real}}$ the viscous damping,
$\tau_{f,j}(\dot{\theta})$ a velocity-dependent friction term (Coulomb friction
and stiction), and $\tau_{\mathrm{ext},j}$ an external load torque.
Linearizing the mechanical dynamics around the operating point and taking the
torque-to-angle relation, we describe each joint by a plant transfer function
$P(s) = \theta_j(s) / \tau_j(s)$. From the inertia and damping terms of
(\ref{eq:sim_dynamics}) and (\ref{eq:real_dynamics}) we obtain
$P_{\mathrm{sim}}(s) = \frac{1}{I_{\mathrm{sim}} s^2 + D s}$, $P_{\mathrm{real}}(s) = \frac{1}{J_{\mathrm{real}} s^2 + d_{\mathrm{real}} s}$,
where the nonlinear and exogenous terms of (\ref{eq:real_dynamics})---the
velocity-dependent friction $\tau_{f,j}(\dot{\theta}_j)$, the disturbance
current $K_{\tau i} i_{d,j}$, and the external load $\tau_{\mathrm{ext},j}$---enter
$P_{\mathrm{real}}$ additively as a lumped disturbance torque.
The \emph{actuator gap} is the multiplicative uncertainty of the real plant
relative to the simulated one,
\begin{equation}
    \Delta P_j(s) = \frac{P_{\mathrm{real}}(s)}{P_{\mathrm{sim}}(s)} - 1
    = \frac{I_{\mathrm{sim}} s + D}{J_{\mathrm{real}} s + d_{\mathrm{real}}} - 1,
    \label{eq:gap}
\end{equation}
so that $P_{\mathrm{real}}(s) = \big(1 + \Delta P_j(s)\big) P_{\mathrm{sim}}(s)$.
This dimensionless gap captures the inertia mismatch
$J_{\mathrm{real}} - I_{\mathrm{sim}}$ and the damping mismatch
$d_{\mathrm{real}} - D$, while the friction, cogging, and external load act as
additive disturbances on top of $\Delta P_j(s)$. It is the same model error
$\Delta P$ that the controller in Sec.~\ref{sec:method} attenuates.
Our objective is to design a controller that suppresses the effect of the actuator
gap, $\Delta P_j \to 0$ and the disturbance influence, so that the closed loop reduces to the simulated
reference dynamics; consequently
$\|\theta_j(t) - \theta_{\mathrm{ref},j}(t)\| \to 0$ for all bounded
disturbances $\tau_{\mathrm{ext},j}$, where $\theta_{\mathrm{ref},j}$ is the
output of the simulator reference model (\ref{eq:sim_dynamics}) driven by the
same policy goal joint angles $\bar{\theta}_j$.
If this objective is achieved, the policy $\pi$ perceives identical
dynamics in simulation and on the real hardware, enabling zero-shot transfer.

\section{Method}
\label{sec:method}

\begin{figure}[t]
  \centering
  \includegraphics[width=0.85\linewidth]{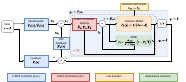}
  \caption{Proposed actuator reality shaping architecture (single joint):
    a cascaded 2-DoF controller (a position outer loop around a velocity inner
    loop) augmented with a disturbance observer (DOB) drives the real motor so
    that its closed-loop response tracks the simulator's reference model.}
  \label{fig:system}
\end{figure}

Figure~\ref{fig:system} shows the proposed per-joint actuator reality shaping
architecture: a cascaded two-degree-of-freedom (2-DoF) controller---a position
outer loop wrapped around a velocity inner loop and augmented with a disturbance
observer (DOB)---that drives the real motor to reproduce the simulator's
reference dynamics. We detail each component in the following subsections.

\textbf{Two-Degree-of-Freedom Controller}
We adopt the standard 2-DoF control
structure~\citep{sugieGeneralSolutionRobust1986,araki2dof1984} to independently specify the reference tracking
and disturbance rejection characteristics.
Let $P(s)$ denote the plant block.
The control input is:
\begin{equation}
    u = \frac{F(s)}{\hat{P}(s)} r + K(s) e, \qquad e = F(s) r - y,
    \label{eq:2dof}
\end{equation}
where $r$ is the reference (the policy goal joint angle $\bar{\theta}$), $y$ is the
measured output ($\theta$), $F(s)$ is the desired system response block, and
$K(s)$ is the feedback block.
Let $P(s) = (1 + \Delta P(s)) \hat{P}(s)$ be the true plant including model
error $\Delta P$ for nominal plant $\hat{P}(s)$.
The closed-loop transfer function from reference to output and from
disturbance to output is:
\begin{equation}
    y = \underbrace{\left(1 + \frac{\Delta P}{1 + P K}\right)}_{\approx 1}
    F r + \underbrace{\frac{P}{1 + P K}}_{\text{sensitivity}} d,
    \label{eq:cl_response}
\end{equation}
where $d$ is the input disturbance torque; the step-by-step derivation is given
in Appendix~\ref{app:2dof_derivation}.
When the feedback gain $K$ is sufficiently large to compensate for the model error ($\Delta P$), the output tracks $y\sim F\,r$
regardless of the disturbance rejection loop, so $F$ can be designed independently
to match the simulator's reference dynamics.
We choose:
\begin{equation}
    F_1(s) = \frac{K_p}{I_\mathrm{sim} s^2 + D s + K_p}, F_2(s) = \frac{\alpha}{s + \alpha},
    \label{eq:feedforward}
\end{equation}
so that the outer-loop plant $P_1 \approx F_2/s$, since the inner velocity loop's closed-loop response closely follows the reference model $F_2$ under the inner-loop 2-DoF velocity control.

The feedback block $K(s)$ must include an integrator to guarantee zero
steady-state tracking error in the presence of constant disturbances, by the
internal model principle.
We use:
\begin{equation}
    K(s) = k_p + \frac{k_i}{s} + k_d s,
    \label{eq:fb}
\end{equation}
a proportional-integral-derivative (PID) controller.

\textbf{Disturbance Observer}
The 2-DoF controller is augmented with a disturbance observer (DOB)~\citep{ohnishiRobustMotion1987,sariyildizAnalysisDisturbanceObserver2015} that
estimates the disturbance torque $\hat{\tau}_d$ from the commanded torque
and the measured velocity through a low-pass filter with cutoff frequency
$\alpha_{\mathrm{dob}}$, and cancels it by subtracting a compensation torque
$ \hat{\tau}_d$ from the commanded current before
it is sent to the driver, attenuating the residual model error $\Delta P$ below
$\alpha_{\mathrm{dob}}$.
The detailed derivation, including the filtered disturbance estimate and the
resulting effective plant, is provided in Appendix~\ref{app:dob}, and the
reconstruction accuracy of the estimated disturbance is reported in
Appendix~\ref{app:dob_acc}.

\textbf{Cascaded Position-Velocity Control}
For position-controlled revolute joints (e.g., arm joints), we cascade two
2-DoF loops.
The inner loop is a velocity control loop with plant
$\hat{P}_2= 1 / (Js + d)$ and reference model matched to the
velocity response $F_2$, where $J$ is the nominal inertia and $d$ is the nominal viscous damping.
The outer loop is a position control loop whose reference model $F_1$ is the
full second-order response~(\ref{eq:sim_dynamics}) and plant $P_1 \sim \hat{P}_1 = F_2 / s$.
The outer loop outputs a goal joint velocity $\bar{\dot{\theta}}$ that serves as
the reference for the inner loop.
Implementation details (per-step procedure and DOB low-pass filter) and the
controller parameters used in all experiments are provided in
Appendix~\ref{app:impl} (Table~\ref{tab:params}).

\section{Experimental Setup}
\label{sec:setup}

We evaluate \emph{actuator reality shaping} on platforms of increasing task
complexity: a {single-joint testbench} (reference-model tracking under
static and dynamic disturbances), a {7-DOF arm} (end-effector reaching
under inter-joint coupling), and a {wheeled-legged robot} and
{humanoid robot} (zero-shot driving and bipedal walking across
embodiments).

\textbf{Hardware Platform.}
We evaluate on three custom-built, in-house platforms: a {$7$-DOF
robotic arm}; a {wheeled-legged robot} ($31$ DOF), a tripod of
four $7$-DOF arms rigidly coupled at a central body, three terminating in wheel
actuators for combined manipulation and locomotion; and a {humanoid
robot} ($22$ DOF) with two $6$-DOF legs, a $2$-DOF waist, and
two $4$-DOF arms. Their IsaacSim models are built from URDF files whose kinematic
trees come directly from the original CAD data.
Because these models lack the implicit parameter tuning of commercial robots and
vendor-supplied simulation assets, the sim-to-real gap on our platforms is
considerably more severe than on off-the-shelf hardware, making them a stringent
testbed for zero-shot transfer.

All single-joint experiments are conducted on a testbench using the same
joint actuator unit deployed on the robots:
a high-gear-ratio servo with a cycloidal gear reducer (Dynamixel YM070, $99{:}1$, ROBOTIS),
so that the identified motor parameters and controller gains transfer directly
to the multi-joint experiments without re-tuning.
The YM070 is used for the single-joint tracking experiments, the robotic arm
reaching task, and the wheeled-legged robot experiments, whereas the
Dynamixel PH54 ($501.9{:}1$, ROBOTIS) is used for the humanoid walking experiment.
The high gear ratio reflects link-side inertia back through $N^2$
($N{=}99$ for the YM070, $N{=}501.9$ for the PH54), so motor-side inertia dominates
the joint-level dynamics by orders of magnitude.
Inter-joint inertial coupling is therefore small relative to the
motor-side inertia and appears at the DOB as a low-frequency bounded
perturbation within the rejection bandwidth, validating the per-joint
decentralized DOB assumption.

\textbf{Simulator and Policy Training.}
The simulator is Isaac Sim~\citep{NVIDIA_Isaac_Sim,makoviychukIsaacGymHigh2021} with joint dynamics
implemented as~(\ref{eq:sim_dynamics}), and policies are trained within the
IsaacLab learning framework~\citep{mittalOrbitUnifiedSimulation2023}.
RL policies are trained using PPO~\citep{schulmanProximalPolicyOptimization2017};
the actor-critic architecture and per-task training settings are summarized in
Appendix~\ref{app:ppo} (Table~\ref{tab:ppo_hparams}).
For the humanoid walking task, we build on the public implementation of
BeyondMimic~\citep{liaoBeyondMimicMotion2025},
replacing only the actuator model and low-level controller with our 2-DoF+DOB design
while keeping the policy architecture, reward terms, and training pipeline unchanged.
The physical-parameter randomization ranges used for the humanoid walking and
wheeled-robot reaching tasks are listed in Appendix~\ref{app:dr}.

\textbf{Comparison Methods.}
We compare against the following baselines, all running at the same $500$\,Hz
driver-level control frequency.
To isolate the DOB's contribution, we report two configurations of our method:
\emph{2-DoF}, the feedforward--feedback controller~(\ref{eq:2dof}) alone, and
\emph{2-DoF+DOB}, the same controller augmented with the disturbance observer
(see Appendix~\ref{app:dob}); elsewhere we refer to the method simply as 2-DoF,
as the DOB is an optional enhancement rather than a separate architecture.

\textbf{PID Cascade}: the servo's factory default controller---a multi-stage
(cascaded) position--velocity--current PID loop running on the drive firmware.
It is a feedback-only, 1-DoF controller with no feedforward path.
This represents the standard servo-controller baseline a practitioner obtains
out of the box without any actuator-aware redesign.

\textbf{Simple PD}: a single-loop PD controller (no feedforward, no DOB, no
cascade) that outputs the motor current command directly from the joint state,$i = k_p (\theta_{\mathrm{ref}} - \theta) - k_d \dot{\theta}$, with gains $k_p, k_d$ manually tuned for stable tracking on the real hardware.

Additionally, for the reaching task we include a learned actuator-modeling
baseline:

\textbf{ASAP}~\citep{heASAPAligningSimulation2025}: a Delta Action
model trained on real-world rollouts of the arm to compensate the
simulation-to-reality dynamics mismatch, deployed on top of the same policy.
Concretely, a Delta Action policy $\pi_\Delta$ is learned so that the deployed
action augments the base policy output,
\begin{equation}
    a = \pi(s) + \pi_\Delta(s),
    \label{eq:asap}
\end{equation}
where $\pi_\Delta$ is fitted to minimize the mismatch between the simulated
transition under $a$ and the observed real-world transition,
i.e. $\min_{\pi_\Delta} \mathbb{E}\big[\lVert f_{\mathrm{sim}}(s, a) -
s'_{\mathrm{real}} \rVert^2\big]$.
Unlike our method, ASAP requires hardware-specific data collection to fit the
Delta Action model.

\section{Results}
\label{sec:results}

\textbf{Single-Axis Tracking.}
We compare the baselines (PID cascade, Simple PD, 2-DoF) with our 2-DoF+DOB on a
YM070 servo, taking the simulator reference dynamics~(\ref{eq:sim_dynamics}) as
the target. The phase portraits of the position error
$\Delta\theta = \theta - \theta_{\mathrm{ref}}$ and velocity error
$\Delta\dot{\theta} = \dot{\theta} - \dot{\theta}_{\mathrm{ref}}$
(Appendix~\ref{app:phase_portraits}, Fig.~\ref{fig:phase_portraits}a) show that
the PID Cascade and Simple PD controllers leave a large sim-to-real gap---their
position loops track the commanded goal angle rather than the reference dynamics,
producing a systematic bias along $\Delta\dot{\theta}$---whereas the 2-DoF and
2-DoF+DOB clouds stay near the origin owing to the feedforward filter $F$, which
reduces phase lag without overshoot. 2-DoF alone leaves a slight position offset,
indicating the DOB further suppresses disturbance torques such as friction.

Over a $20$\,s sinusoidal reference ($20^\circ$ amplitude, $2.5$\,s period;
$5$ runs), the 2-DoF+DOB controller reduces the mean position deviation by
$96.3\%$ and $94.5\%$ relative to the PID Cascade and Simple PD baselines, and the
mean velocity deviation---the cascade's dominant error mode---by $99.1\%$ relative
to the cascade (Table~\ref{tab:dqdv}). The 2-DoF and 2-DoF+DOB results are similar
in single-axis tracking; the DOB's advantage instead becomes pronounced in the
reaching task.

\begin{table}[t]
  \centering
  \small                       
  \setlength{\tabcolsep}{4pt}   
  \caption{Mean $\pm$ std of the deviation from the simulated reference
    trajectory, in joint space ($|\Delta\theta|$, $|\Delta\dot{\theta}|$) for
    single-axis tracking and in task space ($\|\Delta p\|$, $\|\Delta\dot{p}\|$)
    for the reaching task, for each method. Statistics are computed over
    $5$ trials for both the single-axis tracking and the reaching task.}
  \label{tab:dqdv}
  \begin{tabular}{lcccc}
    \toprule
    & \multicolumn{2}{c}{single-axis tracking} & \multicolumn{2}{c}{reaching task} \\
    \cmidrule(lr){2-3} \cmidrule(lr){4-5}
    Method & $|\Delta\theta|$ [rad] & $|\Delta\dot{\theta}|$ [rad/s] & $\|\Delta p\|$ [m] & $\|\Delta \dot{p}\|$ [m/s] \\
    \midrule
    Cascade & $(4.9 \pm 2.7) \times 10^{-2}$ & $3.6 \pm 1.5$ & $(3.7 \pm 3.5) \times 10^{-2}$ & $(6.1 \pm 2.8) \times 10^{-1}$ \\
    PD & $(3.3 \pm 5.8) \times 10^{-2}$ & $(1.7 \pm 7.5) \times 10^{-1}$ & $(4.7 \pm 4.1) \times 10^{-3}$ & $(6.5 \pm 5.8) \times 10^{-2}$ \\
    2DoF & $(2.6 \pm 1.5) \times 10^{-3}$ & $\mathbf{(1.3 \pm 1.4) \times 10^{-2}}$ & $(2.6 \pm 2.0) \times 10^{-3}$ & $(1.0 \pm 0.6) \times 10^{-1}$ \\
    2DoF + DoB & $\mathbf{(1.8 \pm 1.3) \times 10^{-3}}$ & $(3.2 \pm 3.1) \times 10^{-2}$ & $\mathbf{(1.3 \pm 0.9) \times 10^{-3}}$ & $\mathbf{(1.8 \pm 0.7) \times 10^{-2}}$ \\
    \bottomrule
  \end{tabular}
\end{table}

\textbf{Reaching Task.}
The end-effector error on the reaching task, where the $7$-DOF arm tracks a
lemniscate (figure-eight) trajectory, is reported in the phase portrait of
Appendix~\ref{app:phase_portraits} (Fig.~\ref{fig:reaching_phase}); the
policy trained against the ideal reference model~(\ref{eq:sim_dynamics}) in
IsaacSim transfers zero-shot to real hardware via 2-DoF+DOB, and we compare the
deployed real-robot trajectory against the simulated one. Controllers other than
2-DoF track the target poorly, and as before the DOB further improves tracking.

\begin{figure}[t]
  \centering
  \begin{minipage}[t]{0.30\linewidth}
    \centering
    \includegraphics[width=0.9\linewidth]{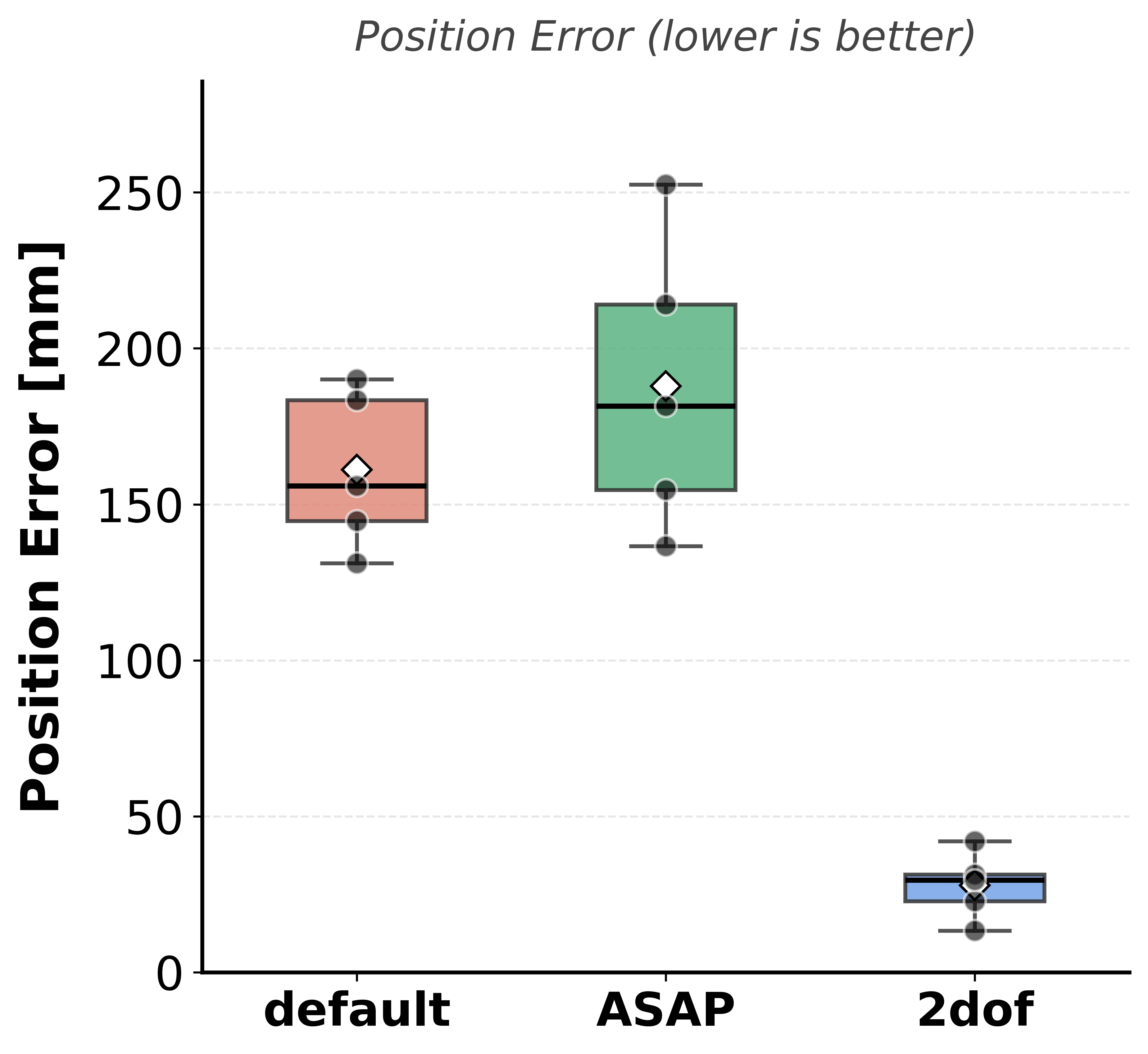}
    \subcaption{End-effector position error}
    \label{fig:asap_pos_err}
  \end{minipage}\hspace{8em}
  \begin{minipage}[t]{0.30\linewidth}
    \centering
    \includegraphics[width=0.9\linewidth]{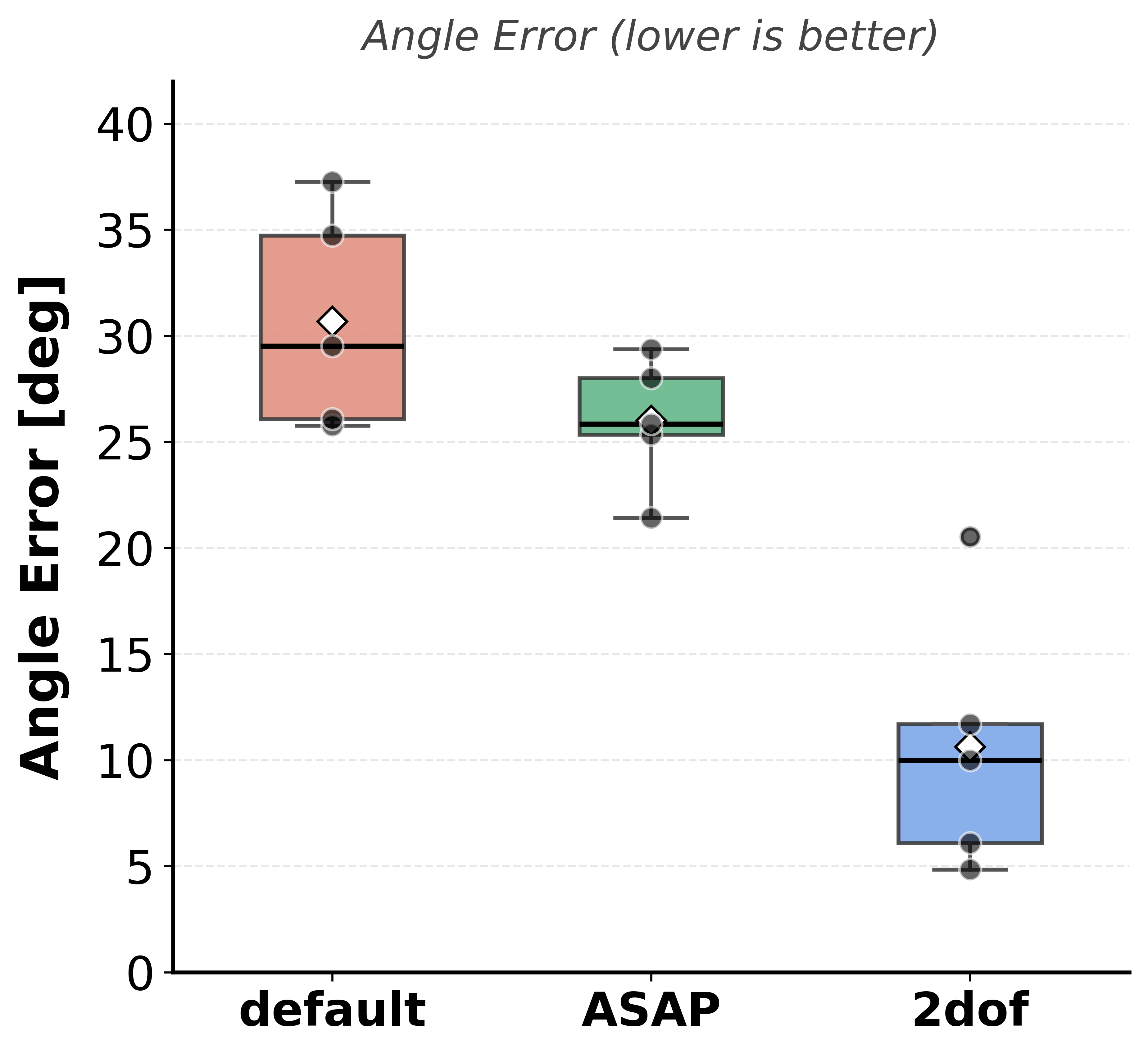}
    \subcaption{End-effector orientation error}
    \label{fig:asap_ang_err}
  \end{minipage}
  \caption{Comparison with ASAP~\citep{heASAPAligningSimulation2025} on the
    reaching task}
  \label{fig:asap_compare}
\end{figure}

Figure~\ref{fig:asap_compare} compares our method against
ASAP~\citep{heASAPAligningSimulation2025}, a real-to-sim-to-real framework that
corrects the simulator from real-world data and retrains the policy.
The ASAP residual~(\ref{eq:asap}) reduced the sim-to-real gap, but to do so it
output larger actions than the base policy $\pi$, destabilizing training in
simulation (loss curves in Appendix~\ref{app:asap_stagea}); its real-robot
performance was thus not substantially better than the untuned factory Default.
Our method instead attains good zero-shot performance without any fine-tuning.

Over the steady-state window of the lemniscate ($5$ runs, Table~\ref{tab:dqdv}),
the 2-DoF+DOB controller attains the lowest end-effector deviation---a $96.5\%$
reduction relative to the factory PID cascade; the DOB accounts for a further
$50\%$ over 2-DoF alone, and 2-DoF+DOB reduces the error by $72\%$ over the
tuned Simple PD baseline, since uncompensated friction in the PD and cascaded
controllers introduces a systematic offset the policy cannot anticipate.

\begin{figure}[t]
  \centering
  \begin{subfigure}[b]{0.8\linewidth}
    \centering
    \includegraphics[width=\linewidth]{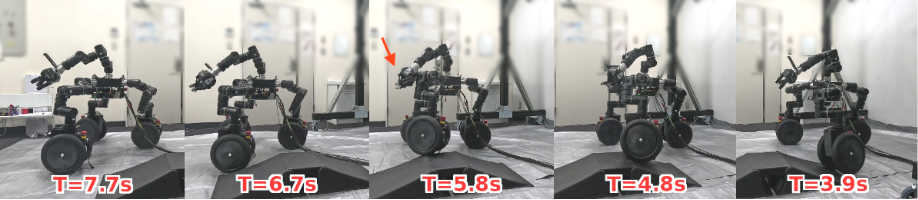}
    \phantomsubcaption
    \label{fig:moonbot_seq}
  \end{subfigure}\\
  \begin{subfigure}[b]{0.8\linewidth}
    \centering
    \includegraphics[width=\linewidth]{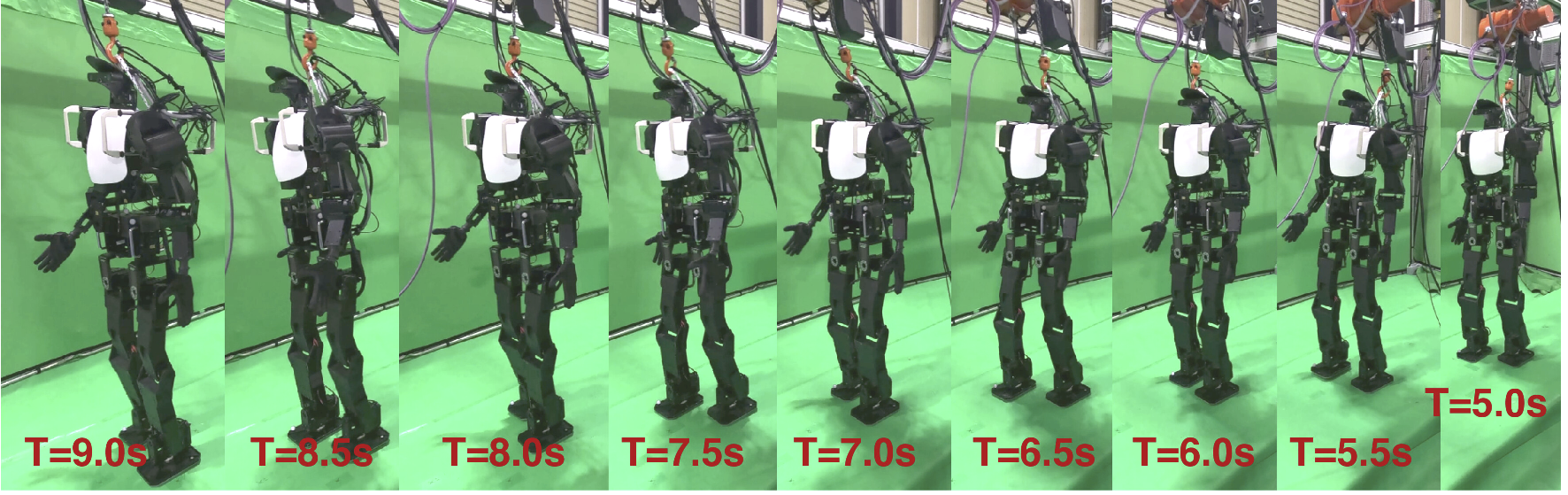}
    \phantomsubcaption
    \label{fig:walk}
  \end{subfigure}
  \caption{\textbf{Zero-shot demonstrations on other embodiments.} The per-joint
    2-DoF driver layer applied to \textbf{(a)}~a wheeled legged robot climbing a
    slope and \textbf{(b)}~a humanoid robot's zero-shot walk.}
  \label{fig:demos}
\end{figure}

\textbf{Zero-Shot Demonstrations on Other Embodiments.}
We further deploy the 2-DoF driver layer on two other platforms; time-lapse sequences are shown in Fig.~\ref{fig:demos}.

\textbf{Wheeled legged robot.} Applied to the velocity-controlled wheel
actuators, a policy trained against the reference model transfers zero-shot,
climbing a slope (Fig.~\ref{fig:moonbot_seq}) whose height must be inferred from
per-joint states and IMU posture alone, difficult for poorly backdrivable
high-gear-ratio servos, yet the 2-DoF controller succeeds by reproducing the
simulator's compliant spring-damper behavior.

\textbf{Humanoid robot.} Using a different servo (PH54), we train a walking policy
with the off-the-shelf BeyondMimic~\citep{liaoBeyondMimicMotion2025} framework,
swapping only the actuator model for the ideal reference model. The policy
transfers without hardware-specific tuning and the robot takes stable forward
steps (Fig.~\ref{fig:walk}), showing the approach carries over to a high-DOF,
dynamically balancing whole-body platform.

\section{Limitations}
\label{sec:discussion}

The method treats the joint-axis inertia as a constant $I_{\mathrm{sim}}$, whereas
it depends on the whole-body configuration $I_{\mathrm{sim}}(q)$; this is
negligible at high gear ratios but not for quasi-direct-drive actuators, where
accounting for it (e.g., via Model Reference Adaptive Control) is a promising
direction. The nominal $J$ and $K_{\tau i}$ still require a coarse estimate, and
whether the approach extends to strongly nonlinear hardware remains open.

\section{Conclusion}
\label{sec:conclusion}
We have presented \emph{actuator reality shaping}: a per-joint
two-degree-of-freedom controller that, using only a minimal motor model, drives
real actuators to replicate the simulator's second-order reference dynamics. By
relocating the sim-to-real adaptation burden to a transparent, hardware-agnostic
driver layer, it enables zero-shot transfer without hardware-specific data.

\acknowledgments{%
 We would like to express our sincere gratitude to Professor Kazuya Yoshida and Assistant Professor Kentaro Uno at Tohoku University and the development team for inspiring the modular robot, MoonBot-mini model. This work was supported by JST ACT-X (Grant No. JPMJAX25CR), JST-Moonshot R\&D Program (Grant No. JPMJMS223B-03), JST-Mirai Program (Grant No. JPMJMI21B1), JST-ASPIRE Program (Grant No. JPMJAP2503), and JSPS KAKENHI (Grant No. 22H04998 and 23K24925) and subsidized by NEDO (Grant No. JPNP14004), Japan.
}
\bibliography{reference}

\appendix

\section{Phase Portraits}
\label{app:phase_portraits}

\begin{figure}[h]
  \centering
  \begin{minipage}[t]{0.45\linewidth}
    \centering
    \includegraphics[width=\linewidth]{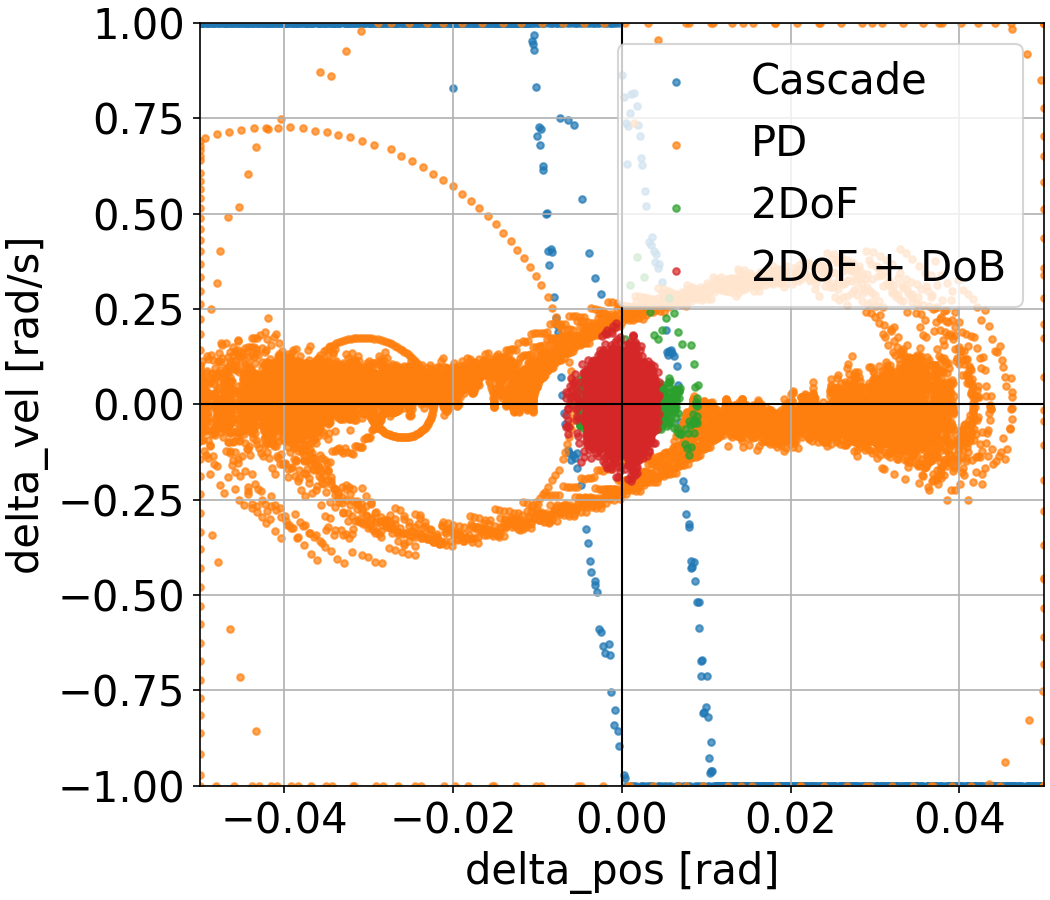}
    \subcaption{Single-axis sinusoidal tracking}
    \label{fig:tracking_phase}
  \end{minipage}\hfill
  \begin{minipage}[t]{0.45\linewidth}
    \centering
    \includegraphics[width=\linewidth]{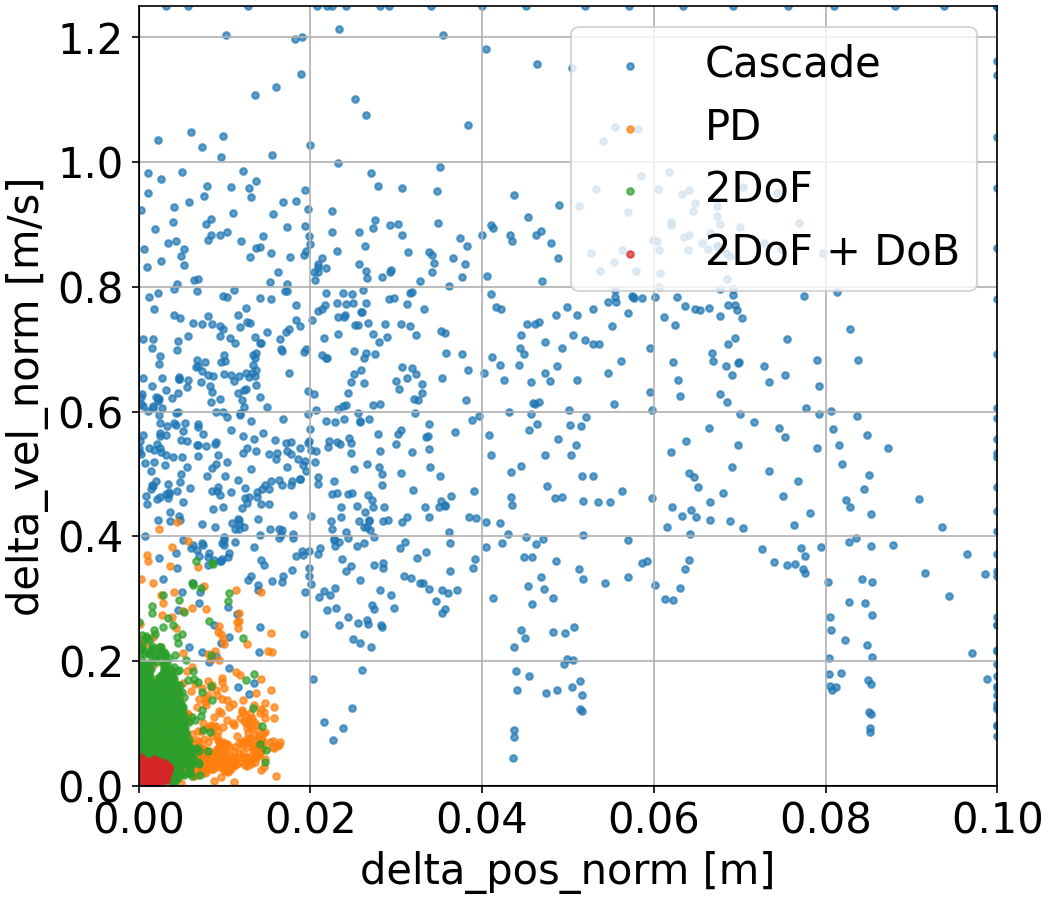}
    \subcaption{Per-joint tracking during reaching}
    \label{fig:reaching_phase}
  \end{minipage}
  \caption{Phase portraits comparing how closely each controller reproduces the
    simulator reference dynamics~(\ref{eq:sim_dynamics}).
    \textbf{(a)}~Single-axis sinusoidal tracking on the YM070 joint
    $(\Delta\theta,\,\Delta\dot{\theta})$, overlaid for all four controllers.
    \textbf{(b)}~7-DoF arm lemniscate reaching, with axes the norms of the
    end-effector position and velocity error vectors. The 2-DoF+DOB cloud
    concentrates near the origin; disabling the DOB or using the PD/cascade
    controllers increases the spread.}
  \label{fig:phase_portraits}
\end{figure}

\section{Derivation of the 2-DoF Closed-Loop Response}
\label{app:2dof_derivation}

This appendix derives equation~(\ref{eq:cl_response}) step by step.
The three governing relations of the 2-DoF structure are
\begin{align}
    y &= P\,(u + d), \label{eq:app_plant}\\
    u &= \frac{F}{\hat{P}}\,r + K\,e, \label{eq:app_input}\\
    e &= F\,r - y, \label{eq:app_error}
\end{align}
where the true plant is related to the nominal model through the
multiplicative uncertainty $\Delta P$ via
\begin{equation}
    P = (1 + \Delta P)\,\hat{P},
    \qquad \text{equivalently} \qquad
    \frac{P}{\hat{P}} = 1 + \Delta P.
    \label{eq:app_uncertainty}
\end{equation}

\paragraph{Step 1: substitute the error into the control input.}
Inserting~(\ref{eq:app_error}) into~(\ref{eq:app_input}) gives
\begin{equation}
    u = \frac{F}{\hat{P}}\,r + K\,(F\,r - y).
    \label{eq:app_step1}
\end{equation}

\paragraph{Step 2: substitute the control input into the plant equation.}
Substituting~(\ref{eq:app_step1}) into~(\ref{eq:app_plant}) yields
\begin{equation}
    y = P\left(\frac{F}{\hat{P}}\,r + K\,(F\,r - y) + d\right)
      = \frac{P}{\hat{P}}\,F\,r + P\,K\,F\,r - P\,K\,y + P\,d.
    \label{eq:app_step2}
\end{equation}

\paragraph{Step 3: collect $y$ on the left-hand side.}
Moving the $P\,K\,y$ term to the left and factoring,
\begin{equation}
    \left(1 + P\,K\right) y
    = \frac{P}{\hat{P}}\,F\,r + P\,K\,F\,r + P\,d.
    \label{eq:app_step3}
\end{equation}

\paragraph{Step 4: apply the uncertainty decomposition.}
Using~(\ref{eq:app_uncertainty}) to replace $P/\hat{P}$ with $1 + \Delta P$,
\begin{equation}
    \left(1 + P\,K\right) y
    = (1 + \Delta P)\,F\,r + P\,K\,F\,r + P\,d
    = \left(1 + P\,K\right) F\,r + \Delta P \, F\,r + P\,d,
    \label{eq:app_step4}
\end{equation}
where the last equality groups the two terms that share the common factor
$F\,r$.

\paragraph{Step 5: divide by the return-difference $1 + P K$.}
Dividing both sides of~(\ref{eq:app_step4}) by $1 + P\,K$ gives the
closed-loop response,
\begin{equation}
    y = \left(1 + \frac{\Delta P}{1 + P\,K}\right) F\,r
        + \frac{P}{1 + P\,K}\,d,
    \label{eq:app_final}
\end{equation}
which is exactly~(\ref{eq:cl_response}).

\paragraph{Interpretation.}
The first term of~(\ref{eq:app_final}) is the reference response: it reduces
to $F\,r$ when the model is exact ($\Delta P = 0$), and any residual model
mismatch is attenuated by the sensitivity function $1/(1 + P\,K)$.
The second term is the input-disturbance response, shaped by the sensitivity
function $P/(1 + P\,K)$.
This decoupling is the design principle exploited in the main text: the
feedforward block $F$ is chosen to match the simulator's reference dynamics,
while the feedback block $K$ is tuned independently to reject disturbances and
steady-state offsets.
Setting $r = 0$ recovers the standard input-sensitivity expression
$y = \frac{P}{1 + PK}d$.
That is, for sufficiently large $K$, the disturbance response is suppressed.

\section{Disturbance Observer}
\label{app:dob}

The DOB estimates the lumped disturbance torque $\tau_d$ from the commanded
torque $\tau = K_{\tau i} i$ and the measured velocity $\dot{\theta}$.
From the motor equation $J \ddot{\theta} = \tau - d \dot{\theta} - \tau_d$,
the disturbance can be extracted as $\tau_d = \tau - (sJ + d) \dot{\theta}$.
Because this expression involves differentiating the velocity signal $\dot{\theta}$,
which amplifies encoder noise, we low-pass filter the estimate:
\begin{equation}
    \hat{\tau}_d = \frac{\alpha_{\mathrm{dob}}}{s + \alpha_{\mathrm{dob}}}
    \left(\tau - (sJ + d) \dot{\theta}\right),
    \label{eq:dob}
\end{equation}
where $\alpha_{\mathrm{dob}}$ is the DOB low-pass cutoff frequency in rad/s.
The estimated disturbance is cancelled by subtracting a compensation current
$i_{\mathrm{comp}} = \hat{\tau}_d / K_{\tau i}$ from the commanded current before
it is sent to the driver.

\section{Disturbance Observer Accuracy}
\label{app:dob_acc}

The nominal motor model ($K_{\tau i}$, $J$, $d$) is estimated from a coarse
one-shot fit of the motor equation~(\ref{eq:real_dynamics}), so a residual model
error $\Delta P$ remains by construction. Since this residual is cancelled at
runtime by the DOB rather than removed offline, we characterize it through the
DOB's reconstruction accuracy below, not through offline identification accuracy.

To evaluate DOB accuracy, we command a sinusoidal target joint angle with
amplitude $20$\,deg and period $2.5$\,s, and compare the torque measured by
the calibrated torque gauge against the torque estimated from the motor
current.
The mean absolute error between the gauge-measured torque and the
current-estimated torque is $0.37$\,N$\cdot$m for YM070 and $0.66$\,N$\cdot$m
for PH54.

\section{Implementation Details}
\label{app:impl}

The 2-DoF controller runs on the motor driver at $f_{\mathrm{ctrl}} = 500$\,Hz.
At each step, the following operations are executed in order:
(1) read encoder angle $\theta$ and compute velocity $\dot{\theta}$;
(2) compute feedforward torque $\tau_{\mathrm{ff}} = \frac{F}{\hat{P}}r$;
(3) compute feedback torque $\tau_{\mathrm{fb}} = K(Fr - \theta)$;
(4) compute DOB estimate $\hat{\tau}_d$;
(5) sum and saturate: $i = \mathrm{clip}((\tau_{\mathrm{ff}} + \tau_{\mathrm{fb}}
- \hat{\tau}_d) / K_{\tau i}, \; -i_{\mathrm{max}}, \; i_{\mathrm{max}})$;
(6) send $i$ to the motor driver.
The velocity $\dot{\theta}$ used by the DOB is passed through an additional
low-pass filter at $0.5 \alpha_{\mathrm{dob}}$ to reduce encoder noise before
entering the DOB numerator, at the cost of a small additional phase lag that is
negligible relative to the control bandwidth.
The controller parameters used in all experiments are summarized in
Table~\ref{tab:params}.

\begin{table}[t]
    \centering
    \caption{Controller parameters used in experiments.
             Hardware parameters ($K_{\tau i}, J, d$) are identified from
             a pendulum free-response fit; controller and simulator
             parameters are design choices.}
    \label{tab:params}
    \begin{tabular}{llrr}
        \toprule
        Parameter & Symbol & PH54 & YM070 \\
        \midrule
        Torque constant       & $K_{\tau i}$ [N$\cdot$m/A]          & $4.8$    & $1.389$ \\
        Nominal inertia       & $J$ [kg$\cdot$m$^2$]                & $0.0763$ & $0.0152$ \\
        Nominal viscous damp. & $d$ [N$\cdot$m$\cdot$s/rad]         & $0.02$   & $1.0$ \\
        Time const. for $F_2$    & $\alpha$ [rad/s]                 & 15      & 30 \\
        DOB cutoff            & $\alpha_{\mathrm{dob}}$ [rad/s]     & 10      & 200 \\
        PID gain for Pos.  & $k_p,k_i,k_d$                          & [15, 0, 0.02]  & [50, 10, 1] \\
        PID gain for Vel.      & $k_p,k_i,k_d$                      & [3,3,0.01]     & [15,8,0.1] \\
        Sim. spring gain      & $K_p$ [N$\cdot$m/rad]               & 50     & 30 \\
        Sim. damping          & $D$ [N$\cdot$m$\cdot$s/rad]         & 3.16      & 3.46 \\
        Sim. inertia          & $I_{\mathrm{sim}}$ [kg$\cdot$m$^2$] & 0.05     & 0.1 \\
        \bottomrule
    \end{tabular}
\end{table}

\section{PPO Training Hyperparameters}
\label{app:ppo}

The actor and critic networks use a standard MLP architecture with ELU
activations, where the number of hidden units per layer is chosen per
task. Table~\ref{tab:ppo_hparams} lists, for each task, the MLP
architecture (hidden units per layer), the number of PPO iterations, the
rollout length per environment, and the number of parallel environments
used during training.

\begin{table}[t]
    \centering
    \caption{PPO training hyperparameters per task.}
    \label{tab:ppo_hparams}
    \begin{tabular}{lcrrr}
        \toprule
        Task & Network (MLP) & Iterations & Rollout/env & Parallel envs \\
        \midrule
        Reaching        & $[512, 256]$      & $4{,}000$  & $128$ & $4{,}096$ \\
        Rover           & $[512, 256, 128]$ & $10{,}000$ & $64$  & $4{,}096$ \\
        Humanoid        & $[512, 256, 128]$ & $30{,}000$ & $24$  & $4{,}096$ \\
        \bottomrule
    \end{tabular}
\end{table}

\section{Domain Randomization}
\label{app:dr}

Tables~\ref{tab:dr_physical} and~\ref{tab:dr_reaching} summarize the
ranges of the physical parameters that are randomized at the start of
each episode for the humanoid walking task and the wheeled-robot
reaching task, respectively. Joint-position resets and observation
noise are omitted from these tables.

\begin{table}[t]
    \centering
    \caption{Domain randomization ranges for physical parameters
    (humanoid walking).}
    \label{tab:dr_physical}
    \small
    \begin{tabular}{llccc}
        \toprule
        Parameter & Target body & Range & Unit & Operation \\
        \midrule
        Static friction   & all bodies & $[0.3,\ 1.6]$       & ---           & uniform   \\
        Dynamic friction  & all bodies & $[0.3,\ 1.2]$       & ---           & uniform   \\
        Restitution       & all bodies & $[0.0,\ 0.5]$       & ---           & uniform   \\
        P gain (stiffness) & all joints & $[80,\ 500]$        & ---           & uniform   \\
        D gain (damping)   & all joints & $[0.8,\ 5.0]$       & ---           & uniform   \\
        CoM offset ($x$)  & torso/base & $[-0.025,\ 0.025]$  & m   & additive  \\
        CoM offset ($y$)  & torso/base & $[-0.05,\ 0.05]$    & m   & additive  \\
        CoM offset ($z$)  & torso/base & $[-0.05,\ 0.05]$    & m   & additive  \\
        \bottomrule
    \end{tabular}
\end{table}

For the wheeled-robot reaching task, actuator gains and joint
parameters are scaled multiplicatively around their nominal values.
The end-effector link mass is scaled to emulate payload variation,
and an external force/torque disturbance is applied to the
end-effector at random intervals to test robustness.

\begin{table}[t]
    \centering
    \caption{Domain randomization ranges for physical parameters
    (wheeled-robot reaching).}
    \label{tab:dr_reaching}
    \small
    \begin{tabular}{llcccl}
        \toprule
        Parameter & Target & Range & Unit & Operation & Distribution \\
        \midrule
        Joint stiffness      & all joints  & $[0.3,\ 3.0]$  & ---       & scale    & log-uniform \\
        Joint damping        & all joints  & $[0.75,\ 1.5]$ & ---       & scale    & log-uniform \\
        Joint friction       & all joints  & $[0.9,\ 1.1]$  & ---       & scale    & uniform     \\
        Joint armature       & all joints  & $[0.9,\ 1.1]$  & ---       & scale    & uniform     \\
        End-effector mass    & ee\_link    & $[0.5,\ 3.0]$  & ---       & scale    & uniform     \\
        External force       & ee\_link\_3 & $[-5,\ 5]$     & N         & additive & uniform     \\
        External torque      & ee\_link\_3 & $[-5,\ 5]$     & N$\cdot$m & additive & uniform     \\
        Disturbance interval & ---         & $[0.5,\ 5.0]$  & s         & ---      & uniform     \\
        \bottomrule
    \end{tabular}
\end{table}

\section{ASAP Model Training Loss}
\label{app:asap_stagea}

To complement the end-effector error comparison in
Fig.~\ref{fig:asap_compare}, we report the training loss for
the delta-action~\citep{heASAPAligningSimulation2025} baseline. ASAP's
delta-action model is fit on real-robot rollouts, and the loss values below
show the per-stage convergence behavior we observed when reproducing this
phase on our hardware. The box plots illustrate that ASAP's correction relies
on hardware-specific data collection, in contrast to the zero-shot transfer
used by our 2-DoF+DOB approach. The model uses a single-step horizon
($H=1$, 1/60 s), predicting the one-step correction from the current state.

\begin{figure}[H]
  \centering
  \includegraphics[width=0.4\linewidth]{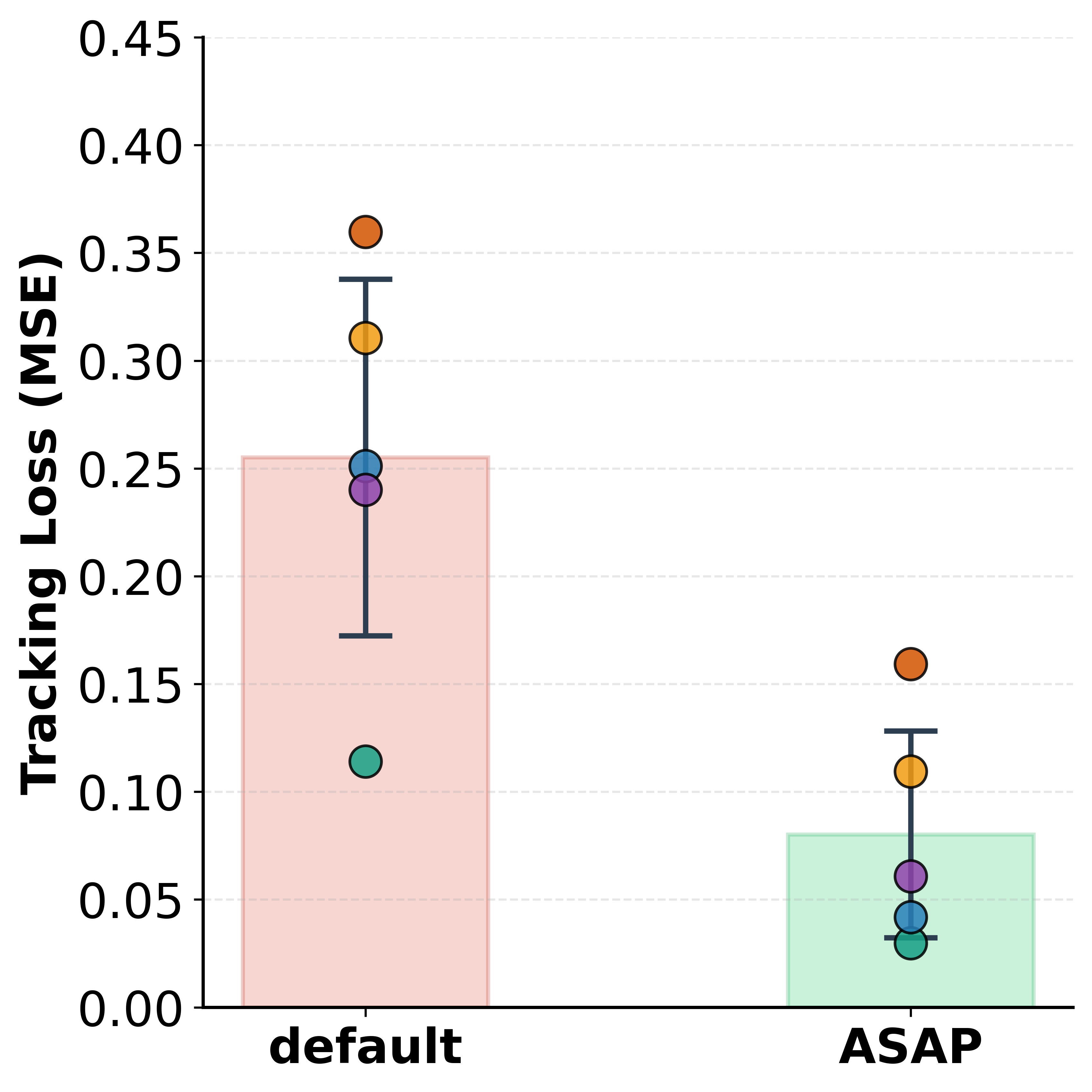}
  \caption{ASAP training-loss (paired runs). Curves correspond
    to the delta-action model fit on real-robot rollouts during ASAP's
    Stage-A phase. Used as supporting evidence for the comparison in
    Fig.~\ref{fig:asap_compare}.}
  \label{fig:asap_stagea_loss}
\end{figure}

\end{document}